\documentclass[runningheads]{llncs}
\usepackage{graphicx}
\usepackage{array}
\newcolumntype{P}[1]{>{\centering\arraybackslash}p{#1}}

\begin{document}

\title{Teach Me What You Want to Play: Learning Variants of Connect Four through Human-Robot Interaction}
\titlerunning{Learning Variants of Connect Four through HRI}
%
\author{Ali Ayub
\and
Alan R. Wagner}
\authorrunning{Ayub, A. and Wagner, A. R.}
%
\institute{The Pennsylvania State University, State College, PA, 16802, USA \\
\email{\{aja5755,alan.r.wagner\}@psu.edu}}
\maketitle              
\begin{abstract}
    This paper investigates the use of game theoretic representations to represent and learn how to play interactive games such as Connect Four. We combine aspects of learning by demonstration, active learning, and game theory allowing a robot to leverage its developing representation of the game to conduct question/answer sessions with a person, thus filling in gaps in its knowledge. The paper demonstrates a method for teaching a robot the win conditions of the game Connect Four and its variants using a single demonstration and a few trial examples with a question and answer session led by the robot. Our results show that the robot can learn arbitrary win conditions for the game with little prior knowledge of the win conditions and then play the game with a human utilizing the learned win conditions. Our experiments also show that some questions are more important for learning the game's win conditions. We believe that this method could be broadly applied to a variety of interactive learning scenarios.\footnote{The final authenticated publication is available online at https://doi.org/10.1007/978-3-030-62056-1\_42} 
    
\end{abstract}

\section{Introduction}
\label{sec:intro}
\noindent The objective of our larger research program is to develop the computational underpinnings and algorithms that will allow a robot to learn how to play an interactive game such as Uno, Monopoly, or Connect Four by interacting with a child. We are motivated by potential applications in hospitals and long-term care facilities for children. Moreover, playing interactive games such as these has been shown to contribute to social development \cite{Buchsbaum12}. Our intent is to create the underlying theory and algorithms that will allow a child to teach a robot to play the games that the child wishes to play. These games may contain nuanced and individualized rules that change and vary each time the game is played or with each child, yet maintain the same underlying basic structure.  

We borrow computational representations from game theory to address this problem. Game theory has been used to formally represent and reason about a number of interactive games such as Snakes and Ladders, Tic-Tac-Toe, and versions of Chess \cite{Berlekamp82}. Game theory offers a collection of mathematical tools and representations that typically examine questions of strategy during an interaction or series of interactions. The term game is used to describe the computational representation of an interaction or series of interactions. Game theory provides a variety of different representations, but the two most common representations are the normal-form game and the extended-form game (described in greater detail below). We use the term "interactive game" to indicate a series of interactions that happen through a board, cards, or play style which has predefined rules, actions, winners and losers. Given this terminology, game theory provides computational representations (games) that can be used to represent interactive games.   

Using representations from game theory has advantages and disadvantages. On the positive side, game theoretic representations have been designed to capture the information needed to formally represent an interaction. Moreover, representing interactions as game-theoretic games allows one to apply the tools and results from game theory as needed \cite{Wagner16}. For example, calculating Nash equilibrium to influence one's play. On the other hand, game-theoretic representations are not easily learned solely from data \cite{Gao12}.

This paper focuses on developing the computational underpinnings necessary for a robot to play the game Connect Four and its variants. In our prior work \cite{Ayub18}, we made some initial progress towards this goal by showing a robot that can learn the four win conditions of Connect Four. This paper focuses on developing formal underpinnings necessary for the robot to not only learn the four win conditions of Connect Four but also its variants. We further analyze our approach in this paper and quantitatively evaluate the importance of different question types for learning the variants of Connect Four. We believe that the methods developed in this paper will also work for other games and hope to show the general applicability of these techniques in future work.

We seek to develop a system that learns how to play the game by asking people questions about the game. We assume that the robot knows what the game pieces are and how to use them. The focus of this paper is thus on the robot learning the win conditions for the game (i.e. how to win). Our approach leverages the robot's developing representation of the game to guide active learning. Specifically, an evolving game tree indicates to the robot the questions that it must ask in order to gain enough knowledge about the structure of the game to be able play it. Often when one person teaches another person how to play a game they begin by explaining how one wins. Our goal is to develop the computational underpinnings that will allow the robot to learn the win conditions well enough to begin playing, even if the full structure of the game has not been learned\footnote{A preliminary version of this paper was accepted at \cite{Ayub_AAAI_20}.}. The main contributions of this paper are:

\begin{enumerate}
    \item A novel approach that utilizes the evolving game-tree representation to ask questions from a user to learn the game's win conditions.
    \item An approach that can be used to learn different win conditions patterns on the Connect Four board in addition to the four win conditions of Connect Four (column, row, diagonal, anti-diagonal).
    \item An experimental analysis that quantifies the importance of different questions for learning various win conditions on the Connect Four board.
\end{enumerate}

\section{Related Work}
\label{sec:related_work}
\noindent The field of artificial intelligence has a long history of developing systems that can play and learn games \cite{Whitehouse13}. 
Recently, significant progress has been made developing systems capable of mastering games such as Chess, Poker and Go using deep reinforcement learning techniques \cite{Xenou19}. While deep reinforcement learning clearly provides a method for learning how to strategically play a game, this approach requires large amounts of training data and is fundamentally non-interactive \cite{Yu18}. Interpersonal game learning, on the other hand, is an interactive process involving limited data and examples, and play must begin before the structure of the game is fully known in order to maintain the other person's attention and interest. Moreover, with children in particular, rules change dynamically in order to make play more favorable and exciting for the child. Data-driven retraining may not be possible or desirable in this situation.

Deep learning-based meta-learning has been proposed as a means for managing the problem of large training time and massive data sets \cite{Yu18}. Although these approaches can learn how to do a task by just watching a single or few demonstrations, the new task has to be very similar to the task that the robot was originally trained on i.e. a robot trained on picking objects will not be able to learn how to place an object. Moreover, the initial meta-learning phase to train the robot on the same task still requires a large amount of data and time. Hence, the problem of using guided interaction with a human to teach the robot a new concept remains unsolved. Although, researchers have investigated using meta-learning on goal-oriented tasks such as visual navigation in novel scenes \cite{Wortsman_2019_CVPR}, to the best of our knowledge, no meta-learning approach exists for learning interactive games by watching just a single demonstration.

Active learning describes the general approach of allowing a machine learner to actively seek information from a human about a particular piece of data in order to improve performance with less training \cite{Settles09,Ayub_2020_CVPR_Workshops,Ayub_IROS_20,Ayub_RoMan_20}. Typically active learning is framed around a supervised learning task involving labeled and unlabeled data. There are a number of different active learning strategies, the membership query strategy being most related to our work \cite{Angluin01}. For this active learning strategy the learner generates queries for a human focused on specific instances of data. One contribution of this paper (further discussed below) is that we leverage the robot's developing game-theoretic representation to assist with the generation of queries directed at the human. In other words, we use the game theory representation to inform the generation of our queries and to contextualize the resulting answers.

\section{Using Game Theory to Represent Interactive Games}
\label{sec:3}
\noindent An interactive game in which players take alternative turns (like Connect Four) can be represented using extensive-form game format \cite{Ayub18}. In Connect Four, players are required to place round game chips into a 7x6 vertical board. This a perfect information game because at each stage both players have complete information about the state of the game, actions taken by the other player and the actions available to the other player in the next stage. At each turn, both players choose a column to place their respective colored chips, hence in each turn a player has a maximum of seven actions available.

\begin{figure}[t]
\centering
\includegraphics[scale=0.39]{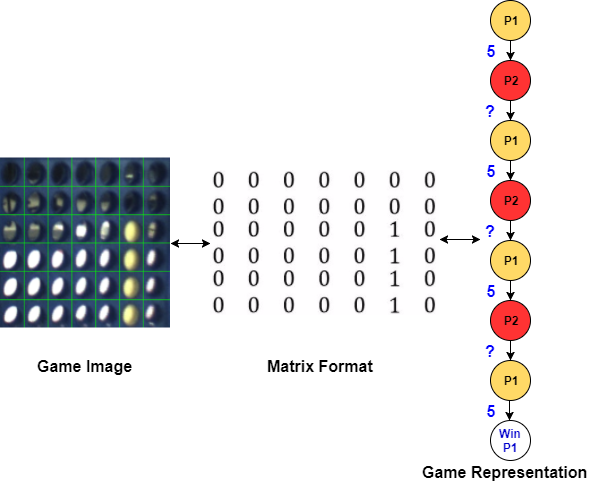}
\caption{A demonstration of column win condition in column 5 for Connect Four seen from the robot’s perspective is shown on the left. The corresponding extensive-form representation is shown on the right. The numbers along with the arrows show the action numbers chosen by the players (5 by the human and ? by the robot since robot’s actions are not shown by the human in the demonstration). Best viewed in color.}
\label{fig:2}
\end{figure}

In order to enable play on a robot, images of a Connect Four game (Fig. \ref{fig:2} left) can be directly translated into a matrix format (Fig. \ref{fig:2} middle) indicating which player has pieces occupying specific positions in the matrix. The matrix format simply encodes the piece positions of the players in the Connect Four board. This matrix can be used to generate an extensive-form game tree (Fig. \ref{fig:2} right). The extensive-form representation can also be translated back into the matrix format and used to generate images of what a game should look like if for an extensive-form representation. The generation of these hypothetical game images afford a method for the robot to communicate with person about possible win conditions (or, more generally, states of the game).

\section{Learning Win Conditions}
\label{sec:4}
\noindent A win condition is a terminal game state in which all players either win, lose or draw. We focus on learning these conditions because doing so is necessary for being able to play the game with purpose. For Connect Four, the rules state that selecting actions that create a pattern of four of the same colored chips in either a row, column, diagonal or anti-diagonal pattern for either player is a win. Players can also draw by filling up the game board without winning. A win condition is represented as a terminal node (a leaf) in the game tree, where one of the players wins the game. All games have some finite set of terminal nodes. The ways to win, lose or draw a game create partitions in the set of terminal nodes based on the game's rules. The purpose of our approach is to learn these general rules through game-theoretic representation of the Connect Four game.

\subsection{Pre-win Condition Learning Tasks}
\noindent Prior to learning a game's win conditions, the robot first needs to capture some basic information about the game structure. In our approach, the robot first asks two questions that allow it to generate a skeleton game structure. The two questions are: "How many players can play this game?" and "Is this a type of game in which players take alternative turns?" The answers to these questions allow the robot to create a generic game tree that simply iterates among the different players. We believe that these questions will be necessary to learn any game. For Connect Four answers to the two questions are "two" and "yes", respectively. 

The robot also needs to know about the components of the game such as the look of the game board, the game chips and their associated colors, and how to physically perform the actions related to the game. We currently assume that this information is pre-programmed and can be loaded once the robot knows the name of the game. We modified the code available at: \url{https://sdk.rethinkrobotics.com/wiki/Connect\_Four\_Demo} for Connect Four which includes the tools for creating the requisite robot behaviors and identifying the game pieces. This pre-programmed information includes:
\begin{itemize} 
  \item[$\bullet$] How to physically perform all of the possible actions \big \{$a_0,a_1,a_2,a_3,a_4,a_5,a_6$\big\} 
  \item [$\bullet$] How to convert a game image into the matrix format of the game state (Figure \ref{fig:2}). 
\end{itemize}

\subsection{From Game Tree to Active Learning}
\noindent From the initial information described above, the robot has as generic game tree structure of the game. The only thing missing from the structure are the win conditions. 
To learn the win conditions of Connect Four and its variants, we use ideas from learning from demonstration and active learning. As a first step the robot asks for a demonstration of a win condition from the human teacher by stating, "Can you please show me a way to win?" It then waits for the person to state, "I am done." The robot converts the image of the board into an extended-form game. For example, Figure \ref{fig:2} depicts the extensive-form representation of a column win in column 5. Note that this demonstration is not the actual game state as it does not depict the red player's moves. However, the robot knows that play iterates between the two players, from the extensive-form representation of Connect Four, so it marks the moves of the red player as unknown.

The initial game tree that exists after the demonstration (Figure \ref{fig:2} right) is clearly missing information. Moreover, the initial tree assumes that player 1 (P1) makes the first move. The demonstration also only depicts a single column win, yet a column win can be achieved in any other column. The demonstration by the human provides only a single game tree branch that leads to a terminal node where P1 wins. Yet there are a large number of other game tree branches that can also lead to terminal nodes. Asking the person to demonstrate each game tree branch that is a win condition is not feasible. The robot thus relies on the extensive-form representation of the game to deduce the information missing from the given demonstration. It then focuses its questions to the human on this missing information, ultimately learning all of the tree branches that could lead to a win condition (terminal node) based on the demonstration. 

From any given demonstration of a win condition (for example Figure \ref{fig:2}), the following information elements are available:
\begin{itemize}
    \item[$\bullet$] \textbf{Given Information:} Winning player's actions. In Figure \ref{fig:2}, these actions are \{5,5,5,5\}
    \item[$\bullet$] \textbf{Missing Information:} The other player's actions and any actions that do not effect the win condition. In Figure \ref{fig:2}, these actions are missing.
    \item[$\bullet$] \textbf{Assumptions:} The robot assumes that P1 takes the first action in the game.
\end{itemize}

\begin{table}[t]
\centering
\small
\begin{tabular}{ |P{3cm}|P{7cm}| }
     \hline
    \textbf{Function Name} & \textbf{Meaning} \\
     \hline
    Translate($Tree$,$p$,$l$) & Change all actions of player $p$ in the $Tree$ by an offset $l$ such that the new actions are between (0-6).\\
     \hline
    AddAction($x$,$p$) & Add an action $x$ for player $p$ \\
    \hline
    RemoveAction($x$,$p$) & Remove an action $x$ for player $p$ \\
 \hline
 \end{tabular}
 \caption{List of functions available to the robot to manipulate the game-theoretic representation of a demonstrated win condition}
 \label{tab:Variables}
 \end{table}

\begin{table}[t]
\centering
\small
\begin{tabular}{ |P{2.4cm}|P{9.5cm}|  }
 \hline
\textbf{Question Type} & \textbf{Example Questions} \\
 \hline
P1 actions  &  Confirm the total number of actions needed by P1 to win the game; Confirm if the actions for P1 can be translated into the game tree. (Definition of Translate in Table \ref{tab:Variables}) \\
 \hline
P2 actions  & What actions can be taken by P2 such that P1 still achieves the win conditions shown in the demonstration?\\
 \hline
Either player's actions  &  What other possible actions can be taken by either player on the game board such that P1 achieves the win condition shown in the demonstration? \\
 \hline
 \end{tabular}
 \caption{\small{The robot asks questions about the winning player's (P1) actions, the losing player's (P2) actions and any other actions taken by the players to learn all the possible win branches that lead to the demonstrated win condition. All these questions are guided by the information elements available from the win condition demonstration and the limited preprogrammed knowledge about the game structure.}}
 \label{tab:QuestionTypes}
\end{table}

Based on the information elements available from the game tree, the robot needs to learn the missing information from the demonstration, confirm the assumptions and learn general rules underlying the given information. These information elements are related to the type of actions that a winning player (P1) and the losing player (P2) can take such that the tree branch leads to a win for P1. Table \ref{tab:QuestionTypes} shows the different questions that the robot needs to ask about both players' actions to learn about the additional information elements about the demonstrated win condition. The questions are pre-programmed in the robot's base knowledge, however when and which questions to ask is guided by the state of the game-tree. Instead of asking the questions verbally (which require a complete dialogue manager), here we present a way for the robot to leverage its ability to convert back and forth between the game state and the game tree. In separate work, we present a dialogue manager than allows the robot to communicate with a human using verbal and visual questions to learn the win conditions \cite{Zare20,Zare19}.

To ask about a specific information element, the robot manipulates the game tree representation of the demonstrated win condition to generate an example situation related to the information that the robot needs to confirm. The robot then converts the manipulated game tree into the game state image and shows it to the human accompanied by a simple yes/no question to confirm whether the example game situation is a win. The simplicity of the question ensures that most people, even older children are capable of providing the robot with an answer. Table \ref{tab:Variables} shows a list of functions available to the robot to manipulate the game tree. These functions are also pre-programmed into the robot's base knowledge.   

Since the robot only asks yes/no questions, it can take multiple example situations for the robot to confirm a single information element. For example, related to the demonstration shown in Figure \ref{fig:2}, to confirm the types of actions P2 can take such that P1 still wins, the robot starts with a general question e.g. can P2 take any actions in the game tree? The answer to that is of course No because if P2 takes action 5 (choose column 5) in its first turn P1 will not achieve a column win in column 5. Hence, the robot asks further clarifying questions to confirm that P2 can take all the actions except the ones that are the same as P1's actions (i.e. action 5) for P1 to achieve a column win. This leads to a hierarchical set of questions asked by the robot, starting with a general to more specific questions. With each more specific question, the robot keeps updating the game-tree representation which guides the next question to be asked. These questions are asked in a visual manner as described above.

The overall flow of our approach for learning the game's win conditions is as follows: The robot starts with a demonstration and continues to ask questions from the human until it confirms all the information elements (Table \ref{tab:QuestionTypes}) needed to learn the demonstrated win condition. This process can also be terminated early if the robot reaches a pre-defined number of questions limit (we set it at 15 questions per win condition for the experiments in this paper).

\begin{figure}[t]
\centering
\includegraphics[scale=0.25]{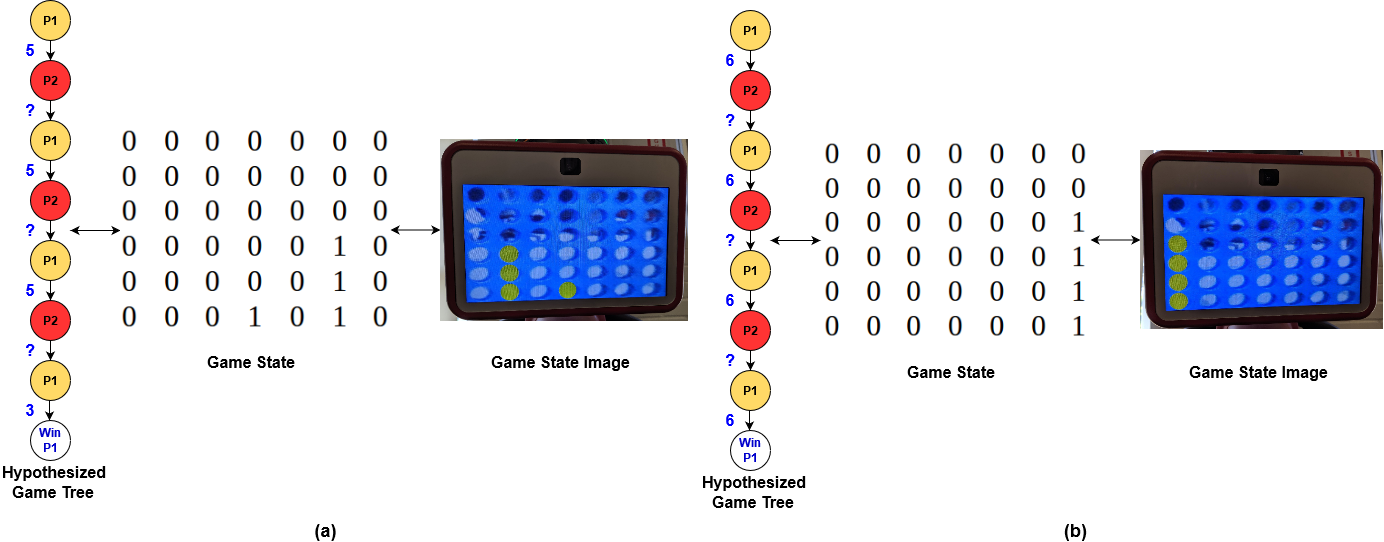}
\caption{(a) The hypothesized game tree generated after changing one action of player 1 in the game tree of Fig. \ref{fig:2} (left). (b) The hypothesized game tree generated after changing all the actions of player 1 to column 6 in the game tree of Fig. \ref{fig:2} (left). For both (a) and (b) the matrix format is from the robot's perspective but the game state image is for the human's perspective. The associated game state image is shown on the right. (Best viewed in color)}
\label{fig:manipulations}
\end{figure}

\noindent \textbf{Example Questions: }To show how the robot asks questions from a human, we show an example session related to one of the questions specific to P1's actions (Confirm if actions for P1 can be translated in the game tree (Table \ref{tab:QuestionTypes})). For this example, we consider the column win demonstration shown in Figure \ref{fig:2}. To learn this information from the human, the robot first confirms if the numerical relationship among all the P1 actions matter i.e. all the P1 actions have to be 5. Since translate operation (in Table \ref{tab:Variables}) is used to change all the actions by a particular offset, a question about translation of all the actions will not be needed if any action can be taken by a player for a win. To confirm this, the robot creates the hypothetical game tree by calling functions \textit{RemoveAction(5,1)} and \textit{AddAction(3,1)} in a sequence to change one of the P1's actions and then converts the manipulated game-theoretic structure to the game-state image (Figure \ref{fig:manipulations} (a)). For the given demonstration, the answer to the accompanied question will be No. Hence, the robot confirms that all the actions of P1 have to be 5. This leads to an update in the game-tree representation as well. Next, using the game-theoretic structure of Connect Four the robot infers that the the siblings of action 5 (columns 0-6 except 5) can also lead to a similar win i.e. P1's actions can be translated in the tree by an offset. To confirm this inference, the robot calls the function \textit{RemoveAction(5,1)} four times to remove all the actions for P1 and then calls the function \textit{AddAction(6,1)} four times to add four actions for P1 in column 6. The manipulated game-theoretic structure is then converted to the game-state image (Figure \ref{fig:manipulations}(b)). The answer to the accompanied question with this example will be Yes for the given demonstration. Hence, the robot confirms an information element about P1's actions in two example situations. Similarly, the robot confirms the other question types from Table \ref{tab:QuestionTypes}.

It should be noted that for board games like Connect Four, the game state can sometimes provide a better representation of a win condition than the game-theoretic structure but the game-state representation is dependent upon a particular game, whereas the game-theoretic structure is completely general. Furthermore, it is easier to reason from the game-theoretic structure than the game-state. Because of this inherent generality of the game-theoretic format to represent any interactive game, our learning algorithm only relies on this representation of interactive games for asking questions and learning about the win conditions. We have shown in related work that the same approach can be used to learn other more complex board games such as Gobblet and Quarto \cite{Zare20}.

\section{Experiments}
\label{sec:experiments}
\noindent To evaluate this system, we used the Baxter robot manufactured by Rethink robotics. Google's text-to-speech API was used to communicate questions in natural language to the person. The person answered the questions by typing inputs into a computer to avoid errors generated by the speech-to-text conversion process. The experimenter served as the robot's interactive partner for all of the experiments, unless stated otherwise.


\subsection{Learning the Four Win Conditions of Connect Four}
We hypothesized that the process described in the previous sections would allow the robot to learn the four Connect Four win conditions (four games pieces in a row, column, or diagonal). We tested the process by providing the robot with a single correct demonstration of one type of win condition (e.g. a column win) and a human then correctly answered the robot’s questions about the self-generated game situations (“Is this a win for yellow?”). We repeated this process for the other types of win conditions (row, diagonal and anti-diagonal). Next, the robot's ability to use the win conditions to play the game was tested in a real game against a human opponent. We verified that the robot could correctly use the win conditions it had learned by playing 10 games against the experimenter. The robot used a depth-2 minimax strategy to play all 10 games. Out of the 10 games, the robot won 7 times, lost 1 and drew 2 times. We believe the reason it lost a game was because it used a depth-2 minimax strategy which only provides the best move for the next stage of the game, not the overall optimal move. Out of the 7 wins, the robot won twice using a diagonal win, 3 times using anti-diagonal and twice using column win. The robot was defeated by a diagonal win in the one game it lost. For all these games, the robot correctly applied the win conditions and demonstrated its ability to correctly identify if it or the person had won the game. These experiments verify that the robot could learn the win conditions from a single demonstration and by using questions and answers to present the person with different game situations, ultimately arriving at a set of extensive-form games constituting a win.

\begin{figure}[t]
\centering
\includegraphics[scale=0.23]{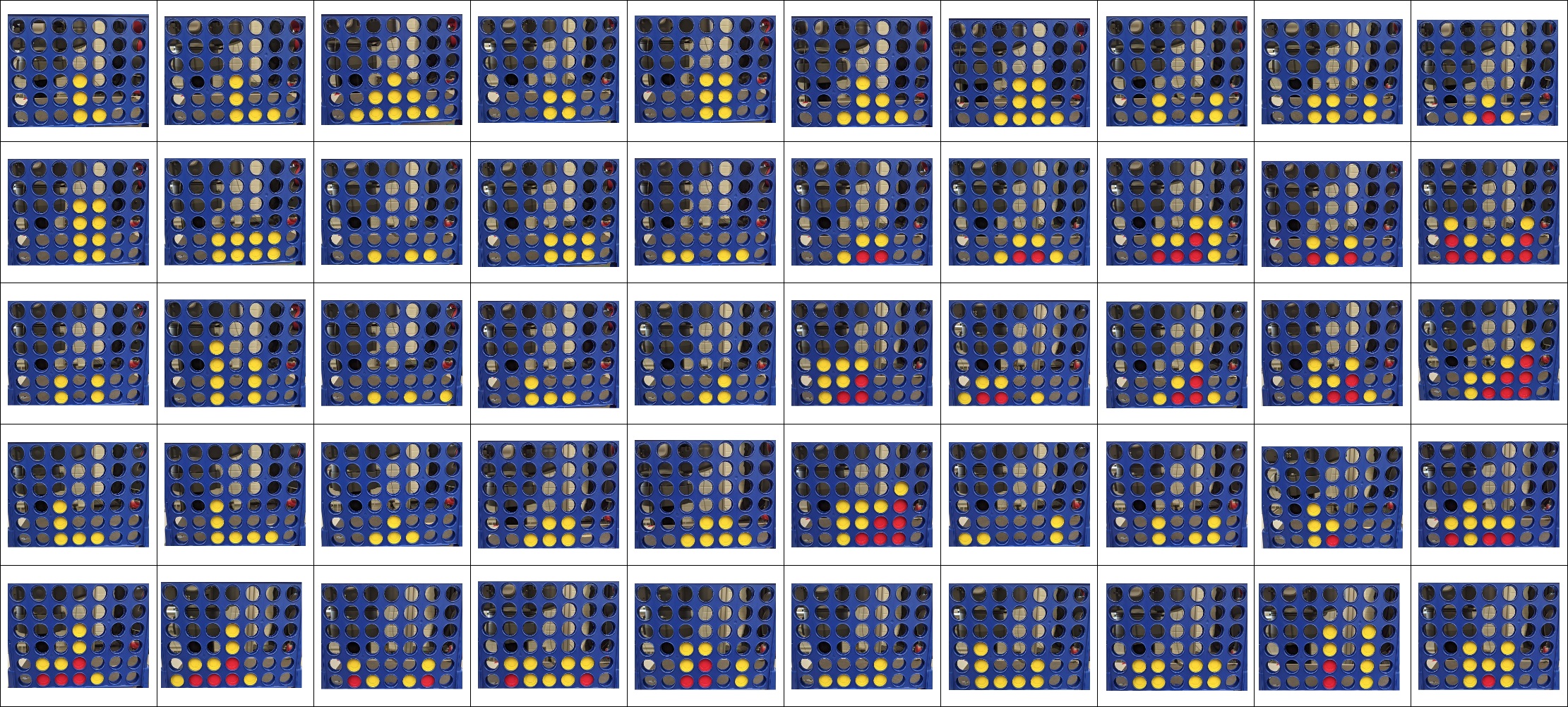}
\caption{Fifty different patterns that were learned by the robot as win conditions on the Connect Four board. Only the yellow chips in the patterns are parts of the win conditions, the red chips are simply to create an offset just like in case of diagonal and anti-diagonal win conditions. Best viewed in color.}
\label{fig:patterns}
\end{figure}

\subsection{Learning Variants of Connect Four}

\noindent To verify that our method is not simply limited to the four win conditions prescribed by the Connect Four game (patterns of four in a row, column, diagonal or anti-diagonal) the robot's ability to learn different patterns representing different ways to win was tested. We hypothesized that our system could learn an arbitrary pattern as a win condition and use this pattern to play a modified version of the game. To test this hypothesis, fifty different randomly generated patterns were demonstrated to the robot as win conditions on the Connect Four game board (Figure \ref{fig:patterns}). The experimenter then answered the corresponding questions for each of the demonstrated win conditions. Once these questions were answered, the robot's ability to use the learned win conditions to play 10 games (for each rule, a total of 500 games) was tested. In these games, both the robot and the experimenter took random actions and all the games ended in an average of 20 turns.  Since the experimenter and the robot both took random actions, instead of checking the robot's ability to play and win using the learned win conditions we simply checked the robot's ability to successfully recognize the learned win condition when it was reached by either the experimenter or the robot. In game-theoretic terms the robot recognizes that a terminal node has been reached. If the robot recognizes a terminal node and ends the game, this implies that the robot has learned the corresponding win condition. Among the 500 games, there were some games (55 games) when the learned win condition was never achieved by the experimenter or the robot and the game ended in a draw. In all 500 games that did not end in a draw, the robot was able to recognize the learned win condition which shows that the robot successfully learned each different win condition on the Connect Four board. We have already shown in the previous experiment if the robot learns a win condition successfully, it can use the minimax strategy to play against a human user. Future user studies will evaluate how well the robot can use the win conditions it has learned to play. This experiment verified the generic ability of our approach to learn various home-made win conditions for a game as long as the structure of the game (board, game pieces, actions available to players in a turn etc.) is known. 

\begin{table}[t]
\centering
\small
\begin{tabular}{ |P{4.3cm}|P{1.2cm}|P{1.3cm}|P{1.3cm}|P{2.1cm}| }
     \hline
    \textbf{Question Type} & \textbf{Row} & \textbf{Column} & \textbf{Diagonal} & \textbf{Anti-diagonal} \\
     \hline
     Min. number of actions for a win & 0\% & 0\% & 0\% & 0\% \\
     \hline
     Translation of P1 actions & 0\% & 0\% & 0\% & 0\% \\
     \hline
     Effect of P2 actions & 13.34\% & 16.67\% & 10\% & 10\%\\
    \hline
    Either Player's Actions & 26.67\% & 90\% & 20\% & 16.67\%\\
 \hline
 \end{tabular}
 \caption{ Detection accuracy (\%) of the robot after removing different question types (from Table \ref{tab:QuestionTypes}) for the four win conditions of Connect Four}
 \label{tab:detection}
 \end{table}
 
\subsection{Importance of Different Question Types}
\noindent For the three question types in Table \ref{tab:Variables}, the robot asks a maximum of 11 questions to learn any win condition pattern on the Connect Four board. Among these 11 questions, a maximum of 4 questions are asked specifically about P2's actions, a maximum of 4 questions are asked about P1's actions (2 for confirming minimum number of actions required for a win and 2 for confirming the translation of P1 actions in the tree) and a maximum of 3 questions are asked about other actions taken by either player in the game. We conducted a final experiment to evaluate the importance of each question type for learning the four win conditions of Connect Four. 

\textbf{Hypothesis:} All three question types are required to learn all the win conditions of Connect Four.

\textbf{Experimental Setup:} The robot learned the four win conditions of Connect Four with one of the question types removed during each interaction. The question type was removed to test the effect of that question type on learning the win conditions. For the questions specific to P1's actions, we further divided them into two groups: to confirm minimum number of actions required for a win and translation of P1's actions. Hence, the robot was taught each win condition in four different interactions and in each interaction one of the question types was not confirmed by the robot (a total of 4*4=16 interactions). After learning each win condition in an interaction, the robot played a total of 30 games with a simulated opponent (total 4$\times$4$\times$30 = 480 games). Both the robot and the opponent took random actions in their turns. 

\textbf{Evaluation:} Since both players took random actions, for each of the games the robot's ability to detect the correct win condition was tested. Table \ref{tab:detection} shows the robot's ability to detect each win condition after removing different question types from the interaction. It is clear that the most important questions are related to P1's actions for all the win conditions. The effect of P2's actions on the win condition learning is also quite drastic. For other actions taken by either player, the column win is least affected (probably because of its simplicity) but all the other win conditions are affected by a significant margin. These results confirm our hypothesis i.e. all question types are necessary for the robot to learn all the win conditions on the Connect Four board but questions specific to P1's actions are the most important.

\section{Conclusion}
\label{sec:conclusion}
\noindent This paper we has demonstrated a method for using game-theoretic representations as a means to structure active learning and incorporate demonstrations in order to learn the win conditions of interactive games. We have presented a preliminary method for using a game tree to generate images of hypothetical game situations that are then presented to a person in order to learn about the game. Our experiments show that a single demonstration accompanied with a few directed questions and answers can be used to learn arbitrary win conditions for the game Connect Four.

We believe, and related work \cite{Zare20} indicates that, the proposed approach can also be used to learn other, more complex games and, perhaps, as a general means for representing interactions between a human and a robot. Ultimately, we believe that this avenue of research may offer a means for a robot to structure its interactions with a person, allowing the robot to bootstrap an interactive exchange by using similar experiences represented as an extended-form game as a model for other upcoming interactions. This paper contributes an important step towards that goal.     

The problem of learning games by interactions with humans is far from solved and the current approach has some limitations. We have assumed that the person demonstrates a valid win condition and that they correctly answer the questions posed by the robot. Further, we did not perform experiments with human participants and only the experimenter interacted with the robots. Future work with human participants will shed more light on the applicability of our approach in real-world scenarios.

This paper suggests several interesting avenues for novel research. Perhaps the most obvious is to extend this work to verbal dialog between a human and the robot. It may be possible to use the game tree to ground open ended answers by the human. This work could also be extended to more completely learn the other aspects of playing a game such as how to perform game actions or use the game components (board, tokens). 
Ultimately, we believe that the proposed techniques take us a step closer to robots that can learn to interact across a wide variety of situations. 

\section*{Acknowledgements}
This work was funded in part by Penn State's Teaching and Learning with Technology (TLT) Fellowship, and an award from Penn State’s Institute for CyberScience.

\bibliographystyle{splncs04}

\bibliography{main}
\end{document}